\documentclass[10pt,twocolumn,letterpaper]{article}

\usepackage{cvpr}              

\usepackage{algorithm}
\usepackage{algorithmic}
\usepackage[accsupp]{axessibility}
\definecolor{cvprblue}{rgb}{0.21,0.49,0.74}
\usepackage[pagebackref,breaklinks,colorlinks,allcolors=cvprblue]{hyperref}

\def\paperID{10286} 
\def\confName{CVPR}
\def\confYear{2025}

\title{Are Spatial-Temporal Graph Convolution Networks for Human Action Recognition Over-Parameterized?}

\author{
Jianyang Xie\textsuperscript{\rm 1}, 
Yitian Zhao\textsuperscript{\rm 2},
Yanda Meng\textsuperscript{\rm 3}, 
He Zhao\textsuperscript{\rm 1},
Anh Nguyen\textsuperscript{\rm 1}, 
Yalin Zheng\textsuperscript{\rm 1}\thanks{Corresponding Author}
\\
\textsuperscript{\rm 1} University of Liverpool, UK. 
\textsuperscript{\rm 2}Ningbo Institute of Materials Technology and Engineering,\\ CAS, China.
\textsuperscript{\rm 3}University of Exeter, UK.\\
{\tt\small \{Jianyang.Xie, yzheng\}@liverpool.ac.uk}
}

\begin{document}
\maketitle
\begin{abstract}
Spatial-temporal graph convolutional networks (ST-GCNs) showcase impressive performance in skeleton-based human action recognition (HAR). However, despite the development of numerous models, their recognition performance does not differ significantly after aligning the input settings. 
With this observation, we hypothesize that ST-GCNs are over-parameterized for HAR, a conjecture subsequently confirmed through experiments employing the lottery ticket hypothesis. Additionally, a novel sparse ST-GCNs generator is proposed, which trains a sparse architecture from a randomly initialized dense network while maintaining comparable performance levels to the dense components. Moreover, we generate multi-level sparsity ST-GCNs by integrating sparse structures at various sparsity levels and demonstrate that the assembled model yields a significant enhancement in HAR performance. Thorough experiments on four datasets, including NTU-RGB+D 60(120), Kinetics-400, and FineGYM, demonstrate that the proposed sparse ST-GCNs can achieve comparable performance to their dense components. 
Even with 95$\%$ fewer parameters, the sparse ST-GCNs exhibit a degradation of $<1\%$ in top-1 accuracy. Meanwhile, the multi-level sparsity ST-GCNs, which require only $66\%$ of the parameters of the dense ST-GCNs, demonstrate an improvement of $>1\%$  in top-1 accuracy. 
The code is available at \url{https://github.com/davelailai/Sparse-ST-GCN}.
\end{abstract}    
\section{Introduction}
\label{sec:intro}

\begin{figure}[ht]
    \centering
    \includegraphics[width=1.0\columnwidth]{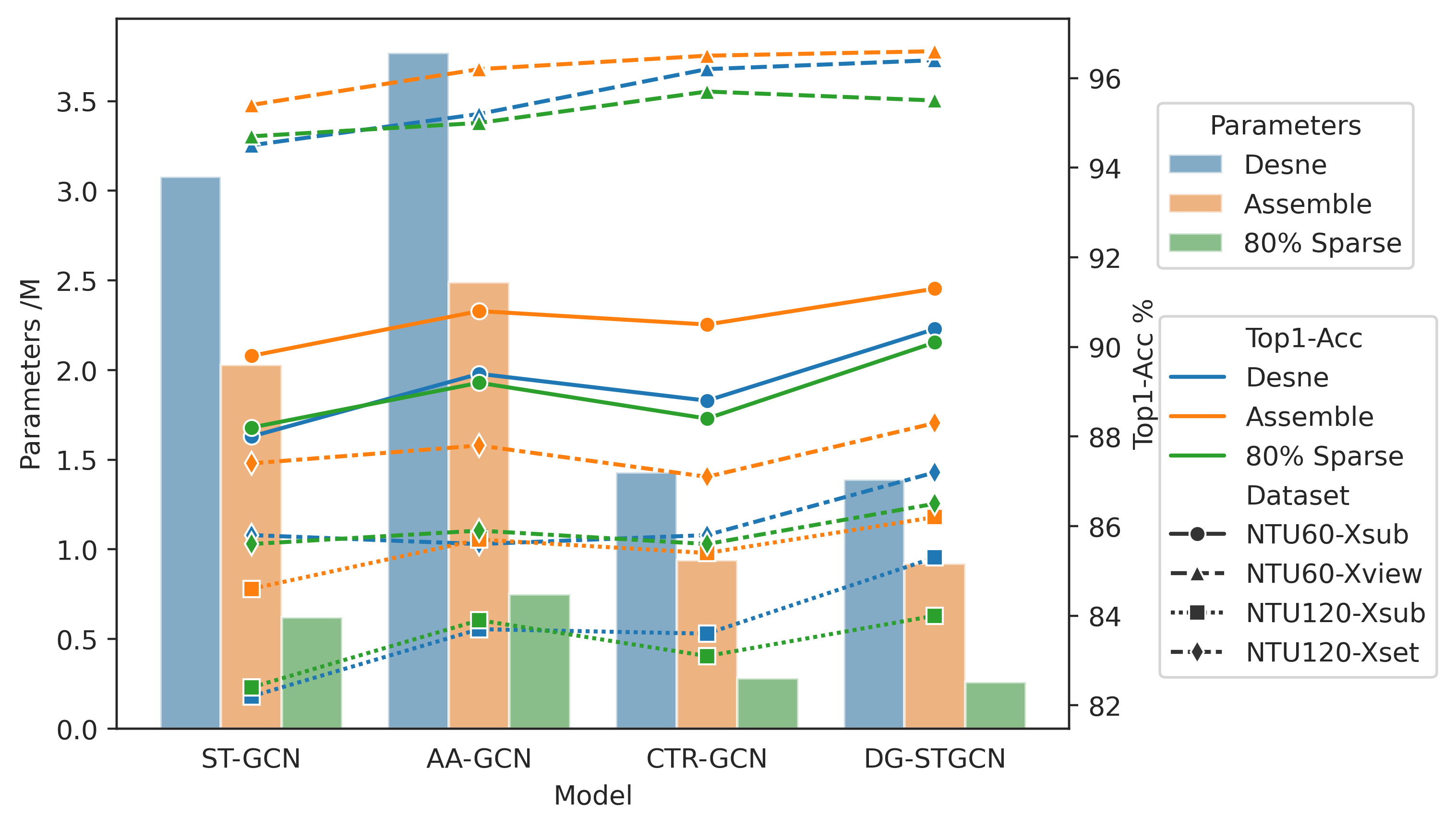}
    \caption{Model comparisons across the NTURGB+D benchmarks. 'Dense' means dense backbone, the deviations of Top-1 Acc across all the four NTURGB+D benchmarks remain below 2$\%$, and the results are independent of the model size. `$80\%$ Sparse' means $80\%$ parameters are masked out, and the sparsity model shows a slight or no degradation in Top-1 Acc when compared with the corresponding dense model. 'Assemble' means multi-level sparsity ST-GCNs by incorporating the sparse structure at different sparsity levels, the results show a significant improvement compared with the corresponding dense model. }
    \label{model_statisic}
\end{figure}
Human action recognition (HAR) is an essential topic in video understanding and has a wide range of applications in intelligent surveillance systems\cite {6126548}, and human-computer interaction\cite{8594452}. Recently, spatial-temporal graph convolution networks (ST-GCNs) \cite{yan2018spatial,si2018skeleton,chen2021channel,cheng2020skeleton,liu2020disentangling,zhang2020semantics,shi2019two} have gained significant popularity. Compared with other methods \cite{vemulapalli2014human,6626306,du2015hierarchical,zhang2017geometric,liu2017skeleton,ke2017new,caetano2019skelemotion,duan2022revisiting} that rely on RGB image sequence \cite{carreira2017quo} or optical flow analysis \cite{simonyan2014two}, the ST-GCNs use body pose and movement information directly, and can effectively capture the interactions between body joints, making it more robust against variations of camera viewpoint and video appearance. 

Although there are many ST-GCN variants with their own merits, their recognition performance does not vary much after aligning the model setting
\cite{duan2022pyskl}. As depicted in Figure \ref{model_statisic}, the deviations of top-1 Acc across all four NTURGB+D benchmarks \cite{shahroudy2016ntu,liu2019ntu} remain below 2$\%$ when considering four different backbones \cite{yan2018spatial,si2018skeleton,chen2021channel,dg2022}, and the result is independent of the model size. On the other hand, the model assembling has demonstrated promising improvements in HAR performance \cite{verma2020deep, das2020vpn,ahn2023star, verma2020deep, kumar2023multi, shi2020skeleton, ding2018combining, ding2023integrating}. However, these approaches typically yield larger models with more parameters, posing challenges for deployment on devices with constrained hardware resources.

Observing these, we hypothesize that: \textit{1) ST-GCNs are potentially over-parameterized for HAR}. \textit{2) Sparse ST-GCNs may enhance the practical application of assembling models by reducing overall model size.} We then substantiated and extended these hypotheses through practical experimentation, and generated a multi-level sparsity ST-GCNs model. This model integrates identical backbones at multiple sparsity levels and demonstrates improved HAR performance without increasing parameters.

Inspired by lottery ticket hypothesis (LTH) \cite{frankle2018lottery,evci2020rigging,zhou2019deconstructing,ramanujan2020s,chijiwa2021pruning}, which states that an independent trainable sub-network exists in a randomly initialized dense model, whose performance is compared to the fully-trained network, we substantiate these hypotheses through practical experimentation in \textbf{Section~\ref{evidence_over_parameterized}}. Our findings reveal that sparse ST-GCNs extracted from fully-trained dense networks achieve comparable performance, even after masking 80$\%$ of the parameters. Additionally, we uncover two notable drawbacks of the sparse ST-GCNs obtained directly from the original LTH \cite{frankle2018lottery}: (1) the performance heavily relies on the initial network weights, and (2) a mask characterized by a high sparse ratio leads to a substantial loss of information, resulting in performance collapse.

Thus, to address the above drawbacks and obtain stable sparse ST-GCNs from random initial weights, we proposed a sparse ST-GCNs generator for the sub-networks extraction from dense ST-GCNs. Firstly, to reduce the reliance on the initial states, the masks in our method are randomly selected or predefined and then kept fixed, while the convolution kernel weights are subject to learning during training. This process enables the conversion of parameters within the randomly selected sub-network into a trainable state. Secondly, to alleviate the loss of information and retain the advantage of the dense network structure, an information compression loss based on the group lasso penalty\cite{lozano2009grouped} was proposed. This penalty loss was utilized to facilitate the information transformation from weights that have been masked to the intended sparse structures.

Furthermore, based on the proposed sparse ST-GCNs generator, we generated multi-level sparsity ST-GCNs by integrating the same models at various sparsity levels. The assembled models have demonstrated a substantial enhancement in HAR performance without parameter increases. This introduces a novel way to improve the overall performance of the assembling ST-GCNs while addressing the challenges of model size. By leveraging sparsity, this approach offers flexibility in model design and implementation, allowing for deployment in diverse scenarios and hardware constraints without compromising efficiency or accuracy.

We summarize our contributions as follows:
\begin{itemize}
\item We successfully substantiated that ST-GCNs exhibit a significant degree of over-parameterization by experiments based on the LTH. To the best of our knowledge, this is the first time to demonstrate a notable degree of over-parameterization in ST-GCNs.

\item We proposed a sparse ST-GCNs generator, which can train stable sparse ST-GCNs from their randomly initialed dense network. Thorough experiments have verified its efficiency across various ST-GCN backbones. The obtained sparse ST-GCNs demonstrate comparable performance to their fully trained dense counterparts.

\item Based on the sparse ST-GCNs generator, we demonstrated that multi-level sparsity ST-GCNs, incorporating backbones at different sparsity levels, enhance HAR performance while reducing parameters, offering a novel approach to optimizing assembling ST-GCNs' overall performance and addressing model size concerns.

\end{itemize}

\section{Related Works}

\subsection{ST-GCNs for skeleton-based HAR}
GCNs have been widely used for skeleton-based HAR \cite{yan2018spatial,si2018skeleton,chen2021channel,cheng2020skeleton,liu2020disentangling,zhang2020semantics,shi2019two,dg2022,xie2024dynamic,xie2024dynamicTIP}.  Yan \etal\cite{yan2018spatial} introduced a pre-defined skeleton graph according to the human body's natural link and proposed the ST-GCN to capture the spatial and temporal patterns from the graph structure. Upon this baseline, some spatial adaptive graph generation methods based on no-local mechanisms were proposed to increase the flexibility of the skeleton graph structure \cite{shi2019two,chen2021channel,cheng2020skeleton,zhang2020semantics, dg2022}.  Instead of only applying the fixed graph structure, these methods learned other adaptive graphs to boost the GCNs' representation ability. For instance, the 2S-AGCN \cite{shi2019two} learned a data-driven adaptive graph for all feature channels, and CTR-GCN\ cite{chen2021channel} learned an adaptive graph for each individual feature channel. Meanwhile, the multi-scale and shift GCN were proposed \cite{cheng2020skeleton,liu2020disentangling} to address the over-smooth problem in graph long-distance transfer. In the temporal pattern, multi-scale temporal convolution was proposed to boost the information aggregation in temporal space \cite{chen2021channel,dg2022}. 

\subsection{Lottery Ticket Hypothesis}
First proposed by Frankle and Carbin \cite{frankle2018lottery}, the lottery ticket hypothesis (LTH) states that there exists a sparse sub-network (called a winning ticket) in a randomly initialized dense network, whose performance is comparable to the fully-trained dense network. Furthermore, several extensions \cite{frankle2020linear, chijiwa2021pruning, ramanujan2020s,zhou2019deconstructing, bai2022dual} have been proposed to enhance its effectiveness. These techniques have observed the existence of untrained sub-networks within randomly initialized convolutional neural networks (CNNs). Remarkably, these sub-networks can achieve comparable accuracy to a fully trained network, even without any updates to the network's weights. Chen \etal\cite{chen2021unified} firstly expended the LTH to the graph content. They introduced a unified GNN sparsification (UGS) framework, demonstrating the effectiveness of LTH across various GNN architectures. While LTH has found applications in various fields \cite{chen2021unified,mikler2022comparing,gan2022playing,chen2020earlybert,chen2020lottery,chen2021ultra}, our work is the first attempt to formulate the LTH in ST-GCNs perspective. Moreover, we extend the LTH to be specifically tailored for ST-GCNs while preserving the advantages of dense network architecture.
\section{Method}
\begin{figure*}[htb!]
    \centering
    \includegraphics[width=1.0\textwidth]{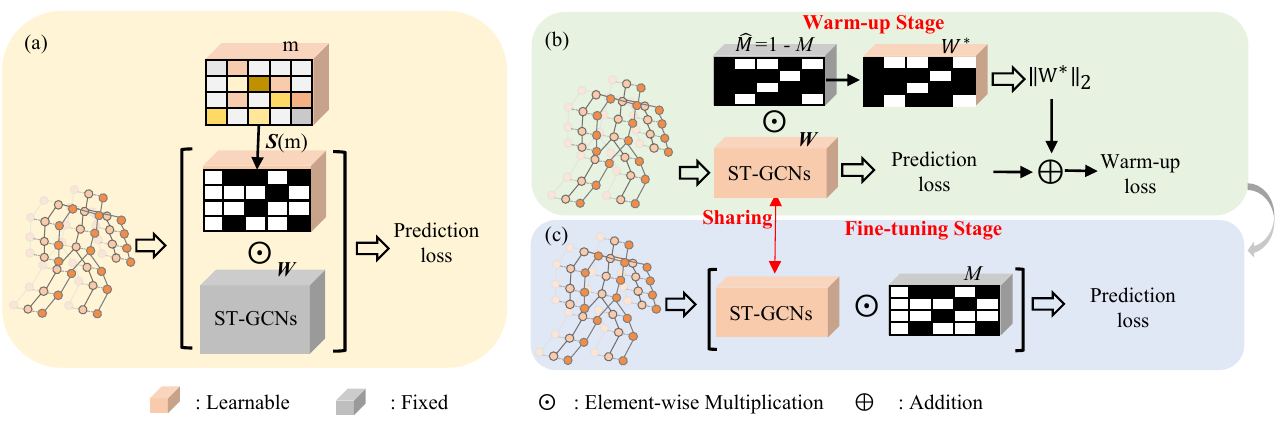}
    \caption{The framework of sparse ST-GCNs generator. (a) represents the Lottery Ticket Perspective of Sparse ST-GCNs, where the weights $W$ in ST-GCNs are fixed, and a learnable mask $m$ is learned. The $S(m)$ represents the binary operation. (b) and (c) represent the two stages of sparse ST-GCNs generator, where the weights $W$ in ST-GCNs are learnable and the mask is pre-defined and kept fixed. During the Warm-up Stage in (b), all the parameters $W$ are involved in training, and the masked parameters $W^*$ with $M=0$ are constrained by penalty loss. During the Fine-Tuning Stage in (c), only the parameters with $M=1$ are engaged in training.} 
    \label{Framework}
\end{figure*}
\subsection{Preliminaries}
\textbf{Notations.} The human skeleton in a video is represented as a spatial-temporal graph and is denoted as $\mathcal{G}=(V, E_s, E_t, X)$, where $V$ is the set of joints, $X$ is the joints feature, $E_s$ and $E_t$ is the spatial and temporal graph respectively. Consider $V=\{v_{ti}| t = 1, ..., T, i = 1,..., N\}$, where $N$ is the number of body joints in each frame, and $T$ is the number of video frames. Here, $v_{ti}$ represents the $n^{th}$ body joints in $t^{th}$ frames. Let $X\in\mathbb{R}^{N \times T \times d}$ represent the joint coordinates as the node feature, where $d$ is the feature dimension. As for $E_s$, an adjacent matrix $A \in \mathbb{R}^{N \times N}$ was defined to describe the overall spatial graph topology, and the edges are formulated as the intro-body connection. The $E_t$ is constructed by connecting the same joints along consecutive frames. ST-GCNs can be divided into two parts: a spatial-GCN (S-GCN) works on the spatial graph $\mathcal{G}_s = (V, E_s, X)$, and a temporal-GCN (T-GCN) works on the $\mathcal{G}_t = (V, E_t, X)$.

\textbf{Spatial-GCN.} The S-GCN works on a spatial graph $\mathcal{G}_s$, and the main operation is to update the node features by aggregating information from its neighborhood within the same frame as Equation~\ref{SGCN}.
\begin{equation} \label{SGCN}
   X_S=\Phi(X, E_s,\theta),
\end{equation}
where $X_S$ represents the outputs of S-GCN, and $\theta$ the parameters of the mapping function. To achieve this, rewriting the input features as  $X = \{X^{t}\in \mathbb{R}^{N  \times d}| t = 1,..., T\}$, and outputs $X_S = \{X_{S}^{t}\in \mathbb{R}^{N  \times C}| t = 1,..., T\}$ of S-GCN can be formulated as Equation~\ref{GCN}, where $C$ is output feature channels.
\begin{equation} \label{GCN}
   X_{S}^{t}=f(AX^t, \theta), t = 1,..., T,
\end{equation}
where $f$ is an updating function with learnable parameters $\theta$. 

\textbf{Temporal-GCN.} The T-GCN works on the $\mathcal{G}_t$, and the key idea is to update the joints' features at the current frame by aggregating the features from its \textit{K-neighbor} frames as Equation ~\ref{TGCN}.
\begin{equation} \label{TGCN}
   X_T=\Psi(X, E_t,\omega),
\end{equation}
where $X_T$ represents the outputs of T-GCN, $\omega$ is the parameters of the mapping function $\Psi$. To achieve this, rewriting $X = \{X^n\in \mathbb{R}^{T  \times d}| n = 1,..., N\}$ as the input feature of T-GCN, the output $X_T = \{X_{T}^{n}\in \mathbb{R}^{T  \times d}| n = 1,..., N\}$ of T-GCN can be formulated as Equation~\ref{TCN}.
\begin{equation} \label{TCN}
   X_{T_{t}^{n}}=\sum_{k=-l}^{l}\omega_k*X_{t+k}^{n}, n = 1,..., N, t = 1,..., T,
\end{equation}
where $X_{T_{t}^{n}}$ represent $n^{th}$ point in the $t^{th}$ frame, $l$ is the window size of temporal convolution; $\omega \in  \mathbb{R}^{2l-1}$ is the learnable weights for feature aggregating.

\textbf{Spatial-Temporal GCN.} A spatial-temporal GCN is generated by operating an S-GCN and a T-GCN sequentially as Equation~\ref{STGCN}.
\begin{equation} \label{STGCN}
   X^{'}=\Psi(X_S, E_t,\omega)=\Psi\big(\Phi(X, E_s,\theta), E_t,\omega\big).
\end{equation}

During the dense ST-GCNs training, let $D=(x,y)$ be the training data, where $x$ is the samples, and $y$ is the corresponding labels. A dense ST-GCN is represented as $\mathcal{F}(x, \theta,\omega)$ with parameters ($\theta$, $\omega$), and $\theta$ and $\omega$ are updated based on the Equation~\ref{denseGCN}.
\begin{equation} \label{denseGCN}
   \theta^{'},\omega^{'}=\mathop{\arg\min}\limits_{\theta, \omega}\mathcal{L}\big(\mathcal{F}(x, \theta,\omega),y\big),
\end{equation}
where $(\theta^{'},\omega^{'})$ is the optimized parameters for the networks and $\mathcal{L}$ is the classification loss function.

\subsection{Lottery Ticket Perspective of Sparse ST-GCNs}
\label{sub_sec_lotteryticket}
As illustrated in Figure~\ref{Framework} (a), a learnable mask $m=(m_s,m_t)$ is introduced, where $m_s$ is the mask for the parameters $\theta$ in S-GCN, and  $m_t$ is the mask for the parameters $\omega$ in T-GCN. The process of sparse ST-GCNs can be formulated as Equation~\ref{sparseSTGCN}.
\begin{equation} \label{sparseSTGCN}
\begin{split}
   X^{'}=\Psi\Big(\Phi\big(X, E_s,\mathcal{S}(m_s)\odot\theta\big), E_t,\mathcal{S}(m_t)\odot\omega\Big)
\end{split},
\end{equation}
where $\odot$ represents element-wise multiplication. $\mathcal{S}$ is the binary operation that controls the sparsity level, $\mathcal{S}(m_s)$ and $\mathcal{S}(m_t)$ represent the binary mask that is utilized for sparse ST-GCNs generation. Thus, the sparse ST-GCNs can be trained following Equation~\ref{sparseGCN_train}
\begin{equation} \label{sparseGCN_train}
\begin{split}
   &m_{s}^{'},m_{t}^{'}=\mathop{\arg\min}\limits_{m_s, m_t}\mathcal{L}\Big(\mathcal{F}\big(x, \mathcal{S}(m_s)\odot\theta,\mathcal{S}(m_t)\odot\omega\big),y\Big)\\
\end{split},
\end{equation}
the threshold of $\mathcal{S}$ is determined as the $S^{th}$ smallest element within $m=[m_s,m_t]$, based on the degree of sparsity.
\subsection{Sparse ST-GCNs Generator}
Despite the feasibility of obtaining sparse ST-GCNs through the algorithm introduced in \textbf{Section~\ref{sub_sec_lotteryticket}}, two drawbacks remain apparent (see \textbf{Section~\ref{evidence_over_parameterized}}): (1) the performance heavily relies on the initial network weight, as the sparse ST-GCNs derived from randomly initialled exhibit considerable performance degradation in comparison to those extracted from pre-trained networks; (2) a mask characterized by a high sparse ratio leads to a loss of information, resulting in performance collapse. 

In order to solve these two limitations, we proposed a sparse ST-GCNs generator. Different from \textbf{Section ~\ref{sub_sec_lotteryticket}}, where the sparse ST-GCN is achieved by training masks for the network's weights, our proposed sparse ST-GCNs generator maintains a static mask while allowing the network weights to be trainable. At the onset of the proposed sparse ST-GCNs generator training, a mask is randomly initialized and then kept constant, while the network's weights are adapted to align with the intended sparse structure. Additionally, to leverage the advantages of the dense architecture, we divided the whole training process into two stages: warm-up and fine-tuning, and proposed an information compression penalty loss, denoted as $L_c$. During the warm-up stage, all network parameters ($\theta$, $\omega$) are optimized and the $L_c$ is incorporated to extract relevant information from the masked portion and then direct it toward the designated sparse network. During the fine-tuning stage, only the parameters selected by the predefined mask are optimized.

\begin{algorithm} 
	\caption{Sparse ST-GCN Generator} 
	\label{alg3} 
	\begin{algorithmic}
		\REQUIRE: $(x,y)$, $M=(M_s,M_t)$, warm-up, epoch, $W=(\theta,\omega)$
            \ENSURE:  $M_{s}$ and $M_{t}$ keep fixed
            \IF{epoch $<$ warm-up}
            \STATE  $W^{*} = W \odot (1-M), $
            \STATE  $\theta^{'},\omega^{'}=\mathop{\arg\min}\limits_{\theta, \omega}\Big(\mathcal{L}\big(F(x, \theta,\omega),y\big)+\lambda\sum_{i=1}^{n}\Vert W^{*}_i \Vert_2\Big)$
            \ELSE
            \STATE $\theta^{'},\omega^{'}=\mathop{\arg\min}\limits_{\theta, \omega}\mathcal{L}\Big(F\big(x, M_s\odot\theta,M_t\odot\omega\big),y\Big)$
            \ENDIF
	\end{algorithmic} 
\end{algorithm}
Specifically, let $M=(M_s,M_t)$ be a randomly selected mask where $M_s$ is the mask for parameters $\theta$ in S-GCN, and  $M_t$ is the mask for parameter $\omega$ in T-GCN, and the parameters masked by $M=0$ as $W^*=(\theta^*,\omega^*)$, and those masked by $M=1$ as $\hat{W^*}=(\hat{\theta^*},\hat{\omega^*})$. During the warm-up stage, $L_c$ is applied to shrink  $W^*$ close to zero, signifying that these parameters have a limited impact on the network and can be eliminated to obtain the final sparse network. Considering there are multi-layers in the whole ST-GCNs structural (normally ten layers), with $W^*=(W^{*}_{i},..., W^{*}_{n})$ where $n$ is the number of ST-GCN layers, a group lasso penalty over the entire set of $W^*$ can be calculated by Equation~\ref{warm_penalty}
\begin{equation} \label{warm_penalty}
L_c= \Vert W^* \Vert_2 = \sum_{i=1}^{n}\Vert W^{*}_i \Vert_2
\end{equation}
Thus during the warm-up stage, the ST-GCNs can be trained following Equation~\ref{warm-up}
\begin{equation} \label{warm-up}
\begin{split}
   & \theta^{'},\omega^{'}=\mathop{\arg\min}\limits_{\theta, \omega}\Big(\mathcal{L}\big(\mathcal{F}(x, \theta,\omega),y\big)+\lambda\sum_{i=1}^{n}\Vert W^{*}_i \Vert_2\Big)\\
\end{split},
\end{equation}
where $\lambda$ is a hyper-parameter to balance the two loss items and set as 1 in our experiments.

During the fine-tuning, only the parameters in the intended sparse structure are involved in training, and the sparse ST-GCNs can be trained by following Equation~\ref{sparseGCN_adapor}.
\begin{equation} \label{sparseGCN_adapor}
\begin{split}
   & \theta^{'},\omega^{'}=\mathop{\arg\min}\limits_{\theta, \omega}\mathcal{L}\Big(\mathcal{F}\big(x, M_s\odot\theta,M_t\odot\omega\big),y\Big)\\
\end{split},
\end{equation}
where $M_s$ and $M_t$ are randomly initiated and kept fixed during the training.
Thus, the overall pipeline of the sparse ST-GCN generator can be illustrated as Algorithm~\ref{alg3}

\section{Experiments}
In this section, joint coordinates were utilized as input features. By extracting the subnet from the fully pre-trained dense ST-GCNs and verifying that the subnet yields comparable results to the original ST-GCNs, we confirmed \textit{\textbf{Hypotheses 1}} regarding the over-parameterization of ST-GCNs. Then we verified the advantages of the proposed sparse ST-GCNs generator on four backbones  \cite{yan2018spatial,si2018skeleton,chen2021channel,duan2022dg} using the NTU-RGBD dataset \cite{shahroudy2016ntu,liu2019ntu}. Finally, we proved that multi-level sparsity ST-GCNs can enhance HAR performance with fewer parameters. Training details can be found in \textit{Supplementary}
\begin{figure*}[tb]
    \centering
    \begin{minipage}[t]{0.6\textwidth}
    \centering
    \includegraphics[width=1\textwidth]{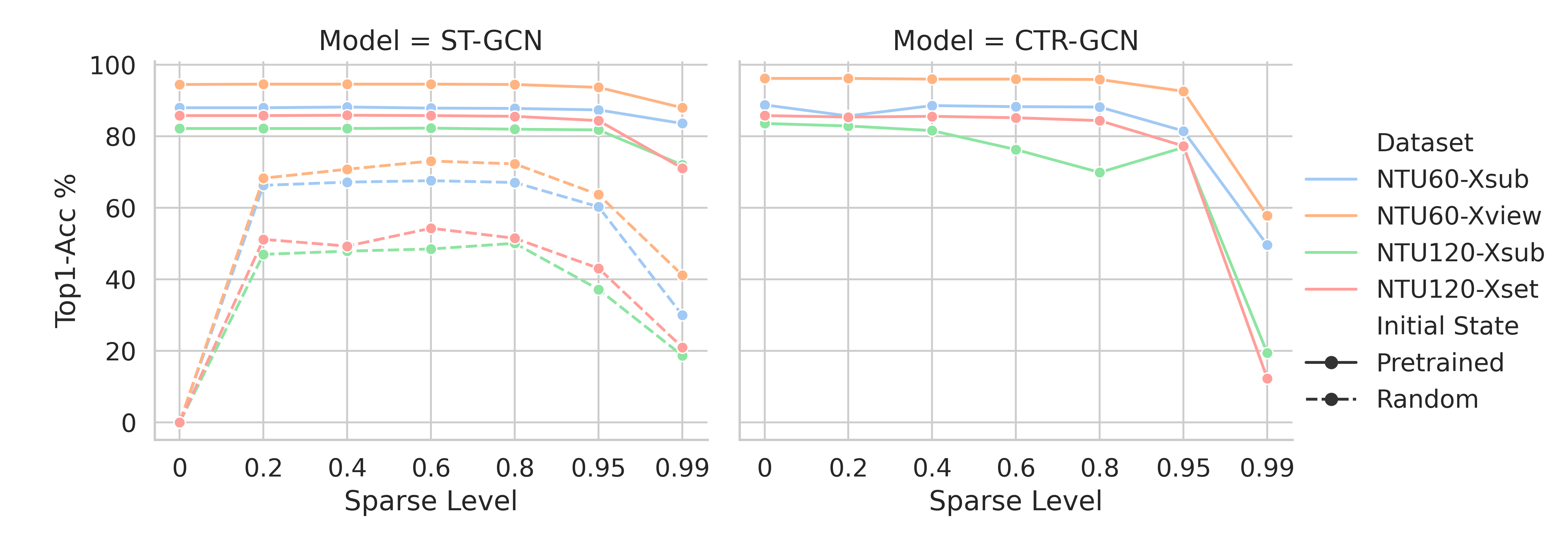}
    \caption{Results for LTH in sparse ST-GCNs. Sparse level means the percentage of masked parameters. 
    The sparse ST-GCNs extracted from pre-trained dense networks can achieve results comparable to the fully trained dense network. 
    which indicates the over-parametrization of the ST-GCNs. 
    However, a notable degradation in the top-1 accuracy is observed at high sparse levels, and the sparse ST-GCN fails to compare favourably with the baseline in the case of the randomly initiated.
    } 
    \label{LTH}
    \end{minipage}
    \quad
    \begin{minipage}[t]{0.35\textwidth}
    \centering
    \includegraphics[width=1\columnwidth]{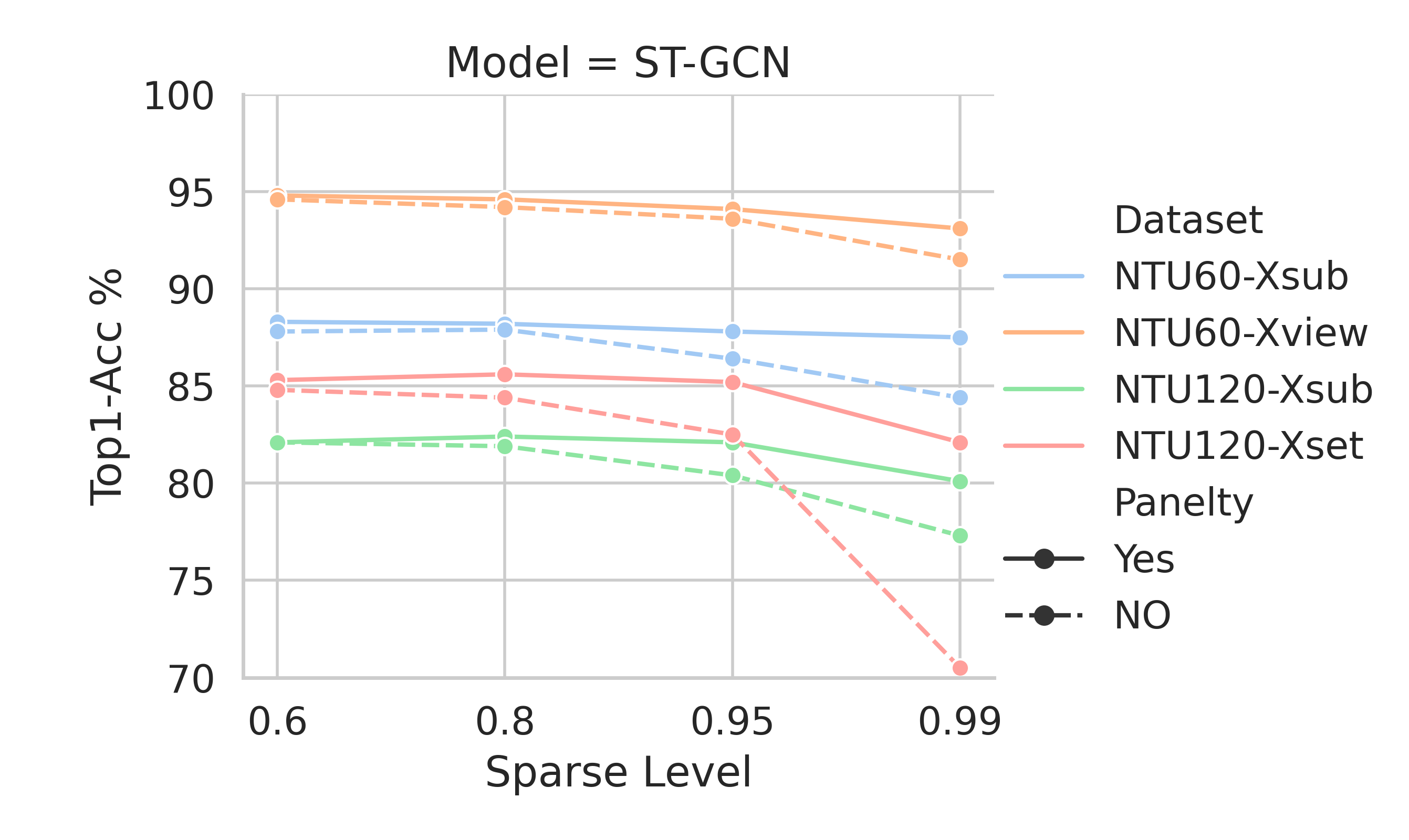}
    \caption{Ablation study of the penalty loss.
    `Yes' means training the warm-up stage with the information compression penalty, and `No' means without. 
    It is obvious that the penalty loss had a positive influence on the performance of sparse ST-GCNs.
    } 
    \label{penalty}
    \end{minipage}
\end{figure*}
\subsection{Datasets}
Four commonly-used datasets were utilized: NTU RGB+D 60 \cite{shahroudy2016ntu}, NTU RGC+D 120 \cite{liu2019ntu}, Kinetics-400 \cite{carreira2017quo}, and FineGYM \cite{shao2020finegym}. The details can be found in \textit{Supplementary}. 

\subsection{Evidence of Over-parameterized in ST-GCNs}
\label{evidence_over_parameterized}
To prove that the dense ST-GCNs are over-parameterized for HAR, we employed the LTH in fully trained ST-GCN \cite{yan2018spatial} and CTR-GCN \cite{chen2021channel}. By extracting the subnet from the pre-trained dense network, we observed that the sparse network achieves a comparable result to the full dense network. As shown in Figure~\ref{LTH}, even when masking out 80$\%$ of the parameters from the pre-trained dense network, there is a slight or no degradation in performance. 

Let's look at the two drawbacks of sparse ST-GCNs based on LTH. From Figure~\ref{LTH}, it can be observed that the ST-GCN with the randomly initiated shows a significant degradation when compared with that pre-trained. On the other hand, As illustrated in Figure~\ref{LTH}, both ST-GCN and CTR-GCN show a significant degradation in performance when the sparsity ratio exceeds 95$\%$. These two drawbacks are effectively tackled by our sparse ST-GCNs generator. 

\subsection{Experiments for Sparse ST-GCNs Generator}
\begin{table*}[tb]
\caption{Performance of the sparse ST-GCNs generator. The names of datasets are represented as NTU60-Xsub(60-sub),  NTU60-Xview (60-view), NTU120-Xsub (120-sub), NTU120-Xset (120-set).}
\begin{center}
\resizebox{\textwidth}{!}{
\begin{tabular}{c|ccccc|ccccc}
\hline
\textbf{Model} &\multicolumn{5}{c|}{\textbf{ST-GCN (Paras:3.10M)}} &\multicolumn{5}{c}{\textbf{AA-GCN (Paras: 3.47M)}} \\
\hline
\textbf{Sparse} & \textbf{Base}  & \textbf{0.6} & \textbf{0.8} & \textbf{0.95} & \textbf{0.99} & \textbf{Base} & \textbf{0.6} & \textbf{0.8} & \textbf{0.95} & \textbf{0.99} \\
\hline
\textbf{60-sub} & 88.0 &\textbf{88.3}(+0.3) &\textbf{88.2}(+0.2) &87.8(-0.2) &87.5(-0.5) & 89.4 &89.3(-0.1) &89.3(-0.1) &\textbf{89.5}(+0.1) &88.6(-0.8)\\
\textbf{60-view} &94.5 &\textbf{94.8}(+0.3) &\textbf{94.6}(+0.1)  &94.1(-0.4) &93.1(-1.4) &95.1 & \textbf{95.4}(+0.4) &95.1(+0.0)  &95.0(-0.1) &94.1(-1.0)\\
\textbf{120-sub} &82.2 & 82.1(-0.1)  &\textbf{82.4}(+0.2) &82.1(-0.1) &80.1(-2.1)  &83.7  &83.0(-0.7) &\textbf{83.9}(+0.2) &83.2(-0.5) &82.2(-1.5)\\
\textbf{120-set} & \textbf{85.8} &85.3(-0.5)  &85.6(-0.2) &85.2(-0.6) &82.1(-3.7) & 85.6  &\textbf{86.0}(+0.4)  &\textbf{85.8}(+0.2) &85.4(-0.2) &84.1(-1.5) \\
\hline 
\hline
\textbf{Model} &\multicolumn{5}{c|}{\textbf{CTR-GCN (Paras: 1.45M)}} &\multicolumn{5}{c}{\textbf{DG-STGCN (Paras: 1.31M)}} \\
\hline
\textbf{Sparse} & \textbf{Base}& \textbf{0.6} & \textbf{0.8} & \textbf{0.95} & \textbf{0.99} & \textbf{Base} & \textbf{0.6} & \textbf{0.8} & \textbf{0.95} & \textbf{0.99} \\
\hline
\textbf{60-sub} & 88.8 &\textbf{89.4}(+0.6) &88.4(-0.4) &88.7(-0.1) &86.7(-1.1) & 90.4  &\textbf{90.5}(+0.1) &90.1(-0.3) &89.4(-1.0) &87.4(-3.0)\\
\textbf{60-view} &\textbf{96.2} & 95.7(-0.5) &95.7(-0.5)  &95.4(-0.8) &92.8(-3.4) &\textbf{96.4} & 95.8(-0.6) &95.5(-0.9)  &94.9(-1.5) &91.4(-5.0)\\
\textbf{120-sub} &\textbf{83.6} & 83.4(-0.2)  &83.1(-0.5) &83.1(-0.5) &79.2(-4.4) &\textbf{85.3} & 84.7(-0.6)  &84.0(-1.3) &83.2(-3.2) &80.7(-4.6)\\
\textbf{120-set} & \textbf{85.8}  &84.9(-0.9)  &85.3(-0.5) &85.2(-0.6) &80.7(-5.1) & \textbf{87.2} &86.1(-1.1)  &86.5(-0.7) &86.2(-1.0) &82.3(-4.9)\\
\hline
\end{tabular}}
\end{center}
\label{St-GCN_div}
\end{table*}
\textbf{Effectiveness of Sparse ST-GCNs Generator.} To validate the effectiveness of the proposed sparse ST-GCNs generator, four different backbones \cite{yan2018spatial,si2018skeleton,chen2021channel,duan2022dg} were employed, considering two datasets \cite{shahroudy2016ntu,liu2019ntu}. The networks were initialized randomly, and sparsity levels of 0.6, 0.8, 0.95, and 0.99 were applied. The performance of the sparse ST-GCNs made by the proposed generator is shown in Table~\ref{St-GCN_div}. Across all four backbones, the sparse network with 80$\%$ of parameters masked out exhibits only a slight or no degradation in performance ($<1\%$).

When evaluating the performance of sparse ST-GCNs under model-size constraints, it is clear that the optimal sparsity level varies for each backbone. As illustrated in Table~\ref{St-GCN_div}, the performance degradation in sparse structures is associated with the model size in a way. Specifically, as the model size increases, the optimal sparsity level also increases. Comparing the AA-GCN (model size: 3.47M) and DG-STGCN (model size: 1.31M), it can be observed that the performance shows a smaller decline in the AA-GCN than in the DG-STGCN when setting the sparse level as the same.  For instance, at 99$\%$ sparsity, AA-GCN's performance degradation in sparse structure is less than 2$\%$, which is lower than DG-STGCN's degradation ($>3\%$).

Comparing the performance of sparse ST-GCNs in each dataset. It is clear that the ideal sparsity level relies on the size of the dataset. As shown in Table~\ref{St-GCN_div}, at the same sparse level,  NTU60 demonstrates less performance degradation compared to NTU120 for all four backbones. Taking the CTR-GCN as an example, when setting the sparse level to 0.99, the performance on NTU60-Xsub experiences a 1.1$\%$ degradation, while both NTU120 witness a degradation of over 4$\%$. This suggests that the optimal sparsity level tends to be lower in larger datasets.

\textbf{Ablation of The Penalty Loss.} In order the validate the influence of the penalty loss in the warm-up stage, the sparse ST-GCN was trained in two settings, with penalty loss and without penalty loss.
The result is shown in Figure~\ref{penalty}, it is evident that the inclusion of the information compression penalty loss in the sparse ST-GCNs generator consistently led to improved performance when compared to training without it. This effect was especially notable at higher sparse levels, where the performance gap became quite substantial. Certainly, considering the performance in NTU120-Xset, the difference in performance between training with the penalty and without it exceeds 10$\%$.
\begin{figure*}[tb]
    \centering
    \includegraphics[width=1.0\textwidth]{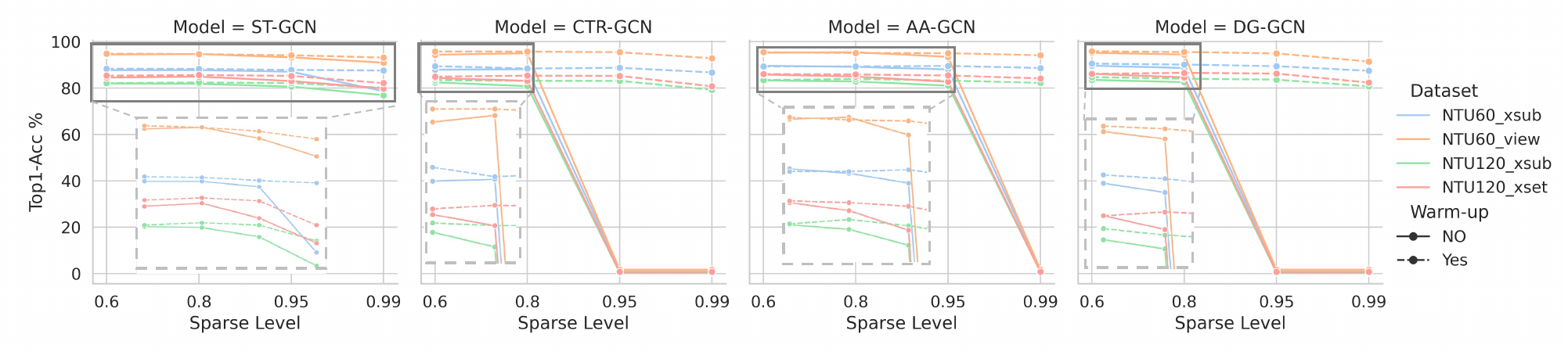}
    \caption{Ablation experiment of the warm-up stage. `Yes' means training the sparse ST-GCNs with the warm-up stage, and `No' means training the sparse ST-GCNs without the warm-up stage, Four backbones were utilized on the NTU RGB+D dataset. 
    }
    \label{warm_up}
\end{figure*}
\textbf{Ablation of The Warm-up Stage.} To verify the importance of the warm-up stage, we trained the sparse ST-GCN generator using two strategies: train the sparse model without a warm-up stage and train the sparse model with a warm-up stage. 
The results are shown in Figure~\ref{warm_up}. It can be observed that the warm-up stage plays an important role in the sparse ST-GCNs generator, particularly at higher sparse levels. At the sparse levels below 0.8, the sparse ST-GCNs generator operates stably even without the warm-up stage, but there is a noticeable degradation in performance when compared to the configurations with the warm-up stage. Meanwhile, as sparsity levels increase significantly, some backbones (AA-GCN, CTR-GCN, and DG-STGCN) without the warm-up stage may suffer performance collapse, while this issue can be mitigated with the warm-up stage. 

Based on the definition in CTR-GCN \cite{chen2021channel}, four backbones can be divided into two types: the graph adaptive-based method (AA-GCN, CTR-GCN, and DG-STGCN) where the skeleton graph is adaptable based on the no-local mechanism, and the graph fixed-based method (ST-GCN), where the skeleton graph is predefined and fixed. 
Considering performance in the context of model type, it is notable that without a warm-up stage, models employing adaptive skeleton graphs are prone to performance collapse under high sparsity levels. For instance, both sparse CTR-GCN and sparse DG-STGCN encounter failure when the sparsity level surpasses 0.8. However, with the warm-up stage, these sparse ST-GCNs achieve results comparable to those of the fully trained dense network, even with 95$\%$ parameters masked out. Meanwhile, in the case of the graph fixed-based method (ST-GCN), the introduction of a warm-up stage resulted in a consistent increase in performance compared to training without a warm-up stage. This effect was particularly pronounced at higher sparse levels, where the performance gap became substantial.

\begin{figure*}[tb]
    \centering
    \includegraphics[width=1\textwidth]{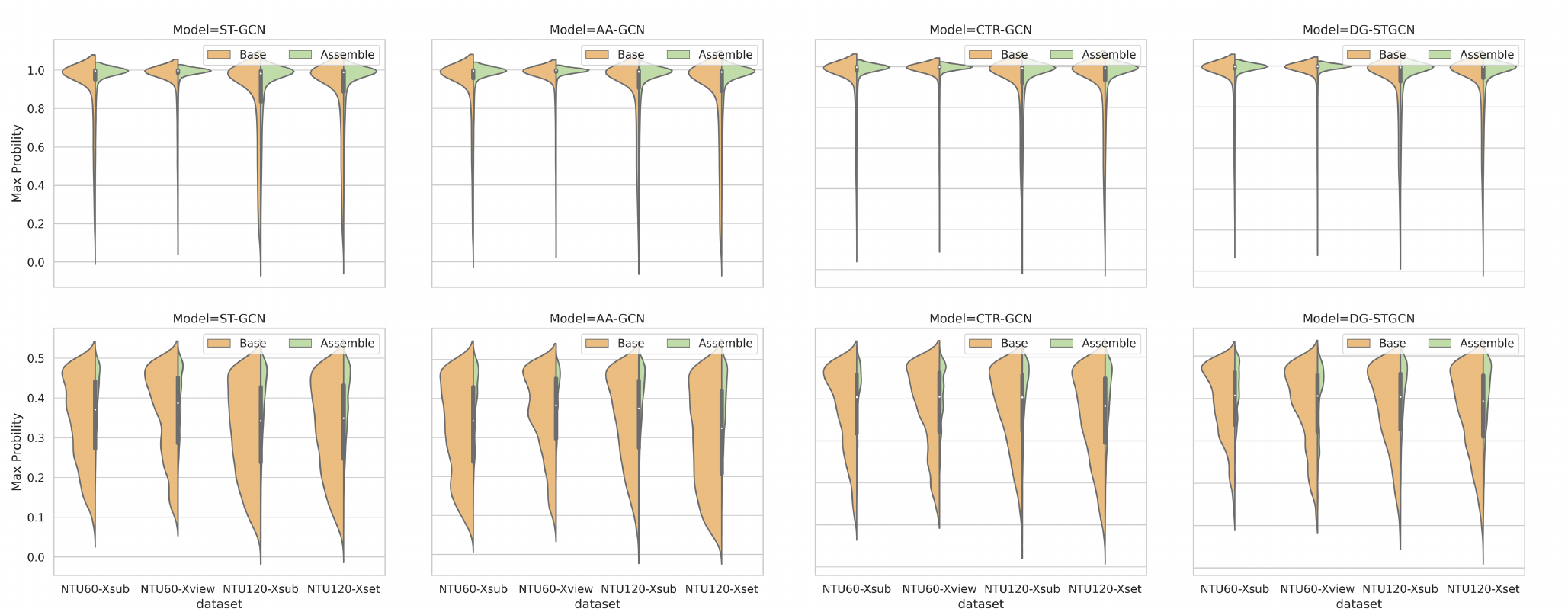}
    \caption{Analysis of the classification distribution for each backbone. 'Base' means the backbone, and 'Assemble' represents the multi-level sparsity structure. The max probability of each sample was utilized as the final result, in the first row, the samples with max probability ranging from 0 to 1 were analyzed. In the second row, the samples with a max probability lower than 0.5 were analyzed. From left to right, the performance of the four backbones is analyzed.}
    \label{assemble}
\end{figure*}
\begin{table*}[htb]
\caption{Experiments for multi-level sparsity ST-GCNs. The `Baseline' is the result of fully trained dense networks, and the `Assemble' is the result of the multi-level sparsity ST-GCNs incorporating the backbones in multiple sparsity levels (0.6, 0.8, 0.95, 0.99 here).}
\begin{center}
\resizebox{0.9\textwidth}{!}{
\begin{tabular}{c|cc|cc|cc|cc}
\hline
\textbf{Module} &\multicolumn{2}{c|}{\textbf{ST-GCN}}  &\multicolumn{2}{c|}{\textbf{AA-GCN}}  &\multicolumn{2}{c|}{\textbf{CTR-GCN}}  &\multicolumn{2}{c}{\textbf{DG-STGCN}} \\
\hline
\textbf{type}&Baseline &Assemble &Baseline &Assemble &Baseline &Assemble &Baseline &Assemble\\
\hline
\textbf{NTU60-Xsub} &88.0 &\textbf{89.8} &89.4 &\textbf{90.8} &88.8 &\textbf{90.5} &90.4 &\textbf{91.3}\\
\textbf{NTU60-Xview}&94.5 &\textbf{95.4} &95.1 &\textbf{96.2} &96.2 &\textbf{96.5} &96.4 &\textbf{96.6}\\
\textbf{NTU120-Xsub}&82.2 &\textbf{84.6} &83.7 &\textbf{85.7} &83.6 &\textbf{85.4} &85.3 &\textbf{86.2}\\
\textbf{NTU120-Xset}&85.8 &\textbf{87.4} &85.6 &\textbf{87.8} &85.8 &\textbf{87.1} &87.2 &\textbf{88.3}\\
\hline
\end{tabular}}
\end{center}
\label{Tab:assemble}
\end{table*}
\subsection{Experiments for Multi-level Sparsity ST-GCNs}
Inspired by the multi-stream fusion framework \cite{shi2020skeleton}, we constructed a multi-level sparsity ST-GCNs incorporating the backbones in multiple sparsity levels. Specifically, we trained four backbones with sparse levels set at 0.6, 0.8, 0.95, and 0.99, and the results for multi-level sparsity ST-GCNs were obtained by aggregating predictions from all sparse structures, as outlined in Table~\ref{Tab:assemble}. It is clear that, for each backbone, the assembled module, yields a significant improvement, but only requires only 66$\%$ of the parameters compared to the dense networks. This presents a novel approach to enhance performance without increasing the model parameters.

To delve into classification performance, we conducted a detailed analysis of the classification distribution across the entire dataset. Utilizing the maximum prediction probability for each sample, rather than the predicted label, we generated violin plots to depict the distribution of classification results. In the first row of Figure~\ref{assemble}, it's evident that the distribution for the proposed assemble models is more concentrated around $1$, indicating the superiority of the multi-level sparsity ST-GCNs. Moreover, to ensure the results' validity, we set a threshold of $0.5$ to define the confidence range. Samples with a maximum prediction probability lower than $0.5$ were then utilized to generate another set of violin plots. In the second row of Figure~\ref{assemble}, we observe a significant reduction in the number of samples with a maximum prediction probability lower than $0.5$ in multi-level sparsity ST-GCNs compared to their backbones. This suggests that the proposed algorithm can confidently classify indistinguishable samples more effectively.

\begin{table*}[tb]
  \caption{
    Comparison with SOTAs. 
    Dataset names are abbreviated as NTU60-Xsub (60-sub), NTU60-Xview (60-view), NTU120-Xsub (120-sub), and NTU120-Xset (120-set).
    DG-STGCN($\mathrm{S}^*$) denotes sparse DG-STGCN with different sparsity levels. 
    DG-STGCN($\mathrm{A}_{0.66}$) represents the multi-level sparsity version.
    Results marked with $^*$ are reported by~\cite{duan2022revisiting}, and those marked with $^o$ are reported by us.
  }
  \label{tab_full_NTU60}
  \centering
  \resizebox{0.87\textwidth}{!}{%
    \begin{tabular}{lcccccc}
      \hline
      \textbf{Module} & \textbf{60-sub} & \textbf{60-view} & \textbf{120-sub} & \textbf{120-set} & \textbf{Kinetics-400} & \textbf{FineGYM} \\
      \hline
      ST-GCN \cite{si2018skeleton}  & 81.5 & 88.3 & 70.7 & 73.2 & 30.7 & 25.2\textsuperscript{*} \\
      SGN \cite{zhang2020semantics} & 86.6 & 93.4 & - & - & - & - \\
      AS-GCN \cite{li2019actional}  & 86.8 & 94.2 & 78.3 & 79.8 & 34.8 & - \\
      RA-GCN \cite{song2020richly}  & 87.3 & 93.6 & 78.3 & 79.8 & 34.8 & - \\
      2s-GCN \cite{shi2019two}      & 88.5 & 95.1 & - & - & - & - \\
      DGNN \cite{shi2019skeleton}   & 89.9 & 96.1 & - & - & - & - \\
      FGCN \cite{yang2021feedback}  & 90.2 & 96.3 & 85.4 & 87.4 & - & - \\
      ShiftGCN \cite{cheng2020skeleton} & 90.7 & 96.5 & 85.9 & 87.6 & - & - \\
      DSTA-Net \cite{shi2020decoupled}  & 91.5 & 96.4 & 86.6 & 89.0 & - & - \\
      MS-G3D \cite{liu2020disentangling} & 91.5 & 96.2 & 86.9 & 88.4 & 38.0 & 92.6\textsuperscript{*} \\
      CTR-GCN \cite{chen2021channel} & 92.4 & 96.8 & 88.9 & 90.6 & - & - \\
      ST-GCN++ \cite{St-gcn++} & 92.6 & 97.4 & 88.6 & 90.8 & \textbf{49.1} & - \\
      PoseConv3D \cite{duan2022revisiting} & \textbf{94.1} & 97.1 & 86.9 & 90.3 & 47.7 & 94.3 \\
      SKE2GRID \cite{cai2023ske2grid} & 93.8 & \textbf{98.6} & 87.3 & 90.8 & - & - \\
      DG-STGCN \cite{dg2022} & 93.2 & 97.5 & 89.6 & 91.3 & 40.3 & 95.1\textsuperscript{o} \\
      \hline \hline
      DG-STGCN($\mathrm{S}_{0.6}$) & 92.7 & 97.3 & 89.1 & 90.8 & 48.4 & \textbf{95.3} \\
      DG-STGCN($\mathrm{S}_{0.8}$) & 92.8 & 97.3 & 89.0 & 90.7 & 47.4 & 95.1 \\
      DG-STGCN($\mathrm{S}_{0.95}$) & 92.8 & 97.1 & 88.7 & 90.4 & 46.8 & 94.8 \\
      DG-STGCN($\mathrm{S}_{0.99}$) & 90.5 & 94.9 & 85.3 & 90.4 & 42.0 & 92.2 \\
      DG-STGCN($\mathrm{A}_{0.66}$) & 92.9 & 97.4 & \textbf{90.4} & \textbf{91.4} & 47.5 & \textbf{95.3} \\
      \hline
    \end{tabular}%
  }
\end{table*}
\subsection{Comparisons with the State-of-the-art}
To ascertain the effectiveness of the sparse ST-GCNs generator within the multi-modalities fusion framework \cite{shi2020skeleton}, we employed the DG-STGCN \cite{dg2022} as the backbone and set the sparse level as 0.6, 0.8, 0.95 and 0.99, the sparse models are represented as DG-STGCN(S*). Following this, we trained the sparse DG-STGCN on four distinct input modalities: joints (j), joint motion (jm), bone (b), and bone motion (bm). The final result was obtained by aggregating the predictions from all streams. The performance of the sparse DG-STGCN was compared with state-of-the-art methods on NTURGB+D \cite{shahroudy2016ntu, liu2019ntu}, Kinetics-400 \cite{carreira2017quo}, and FineGYM \cite{shao2020finegym} in Table~\ref{tab_full_NTU60}. It can be found that the sparse DG-STGCN can obtain a comparable result to the SOTAs.
The multi-level sparsity DG-STGCN (represented as DG-STGCN(A)), which requires only $66\%$ of the parameters of the dense DG-STGCN, obtained the SOTA in 3 datasets and has comparable results in the other three.

\section{Discussion} 
In this paper, we empirically demonstrated the over-parameterization of ST-GCNs in skeleton-based HAR. We showed that sub-networks extracted from fully trained dense ST-GCNs perform comparably to their dense counterparts. Leveraging this insight, we proposed a sparse ST-GCNs generator to learn sparse architectures from randomly initialized dense networks. The generator allows for predefined or randomly initiated sparse structures. Extensive ablation experiments across various backbones confirmed the generalization of our approach, with sparse ST-GCNs achieving comparable performance to dense components, even at high sparsity levels. Additionally, we generated multi-level sparsity ST-GCNs by combining backbones at different sparsity levels, resulting in significant HAR performance improvements with fewer parameters. Notably, our generic generator offers a novel approach for future model applications. 
However, there are avenues for future work. On one hand, the sparsity level is manually set in the current method. Thus, it's imperative to develop mechanisms for automatically learning an optimal sparsity level for various structures. On the other hand, the multi-level sparsity model is currently generated by assembling the model after single-model training. In the future, exploring an end-to-end training approach for multi-level sparsity models is warranted.

\section{Acknowledgment}
This work is supported by the EPSRC (Engineering and Physical Sciences Research Council) Centre for Doctoral Training in Distributed Algorithms (Grant Ref: EP/S023445/1).

{
    \small
    \bibliographystyle{ieeenat_fullname}
    \bibliography{main}

\begin{thebibliography}{55}
\providecommand{\natexlab}[1]{#1}
\providecommand{\url}[1]{\texttt{#1}}
\expandafter\ifx\csname urlstyle\endcsname\relax
  \providecommand{\doi}[1]{doi: #1}\else
  \providecommand{\doi}{doi: \begingroup \urlstyle{rm}\Url}\fi

\bibitem[Ahn et~al.(2023)Ahn, Kim, Hong, and Ko]{ahn2023star}
Dasom Ahn, Sangwon Kim, Hyunsu Hong, and Byoung~Chul Ko.
\newblock Star-transformer: A spatio-temporal cross attention transformer for human action recognition.
\newblock In \emph{Proceedings of the IEEE/CVF winter conference on applications of computer vision}, pages 3330--3339, 2023.

\bibitem[Bai et~al.(2022)Bai, Wang, Tao, Li, and Fu]{bai2022dual}
Yue Bai, Huan Wang, Zhiqiang Tao, Kunpeng Li, and Yun Fu.
\newblock Dual lottery ticket hypothesis.
\newblock \emph{arXiv preprint arXiv:2203.04248}, 2022.

\bibitem[Caetano et~al.(2019)Caetano, Sena, Br{\'e}mond, Dos~Santos, and Schwartz]{caetano2019skelemotion}
Carlos Caetano, Jessica Sena, Fran{\c{c}}ois Br{\'e}mond, Jefersson~A Dos~Santos, and William~Robson Schwartz.
\newblock Skelemotion: A new representation of skeleton joint sequences based on motion information for 3d action recognition.
\newblock In \emph{2019 16th IEEE international conference on advanced video and signal based surveillance (AVSS)}, pages 1--8. IEEE, 2019.

\bibitem[Cai.et.al(2023)]{cai2023ske2grid}
Cai.et.al.
\newblock Ske2grid.
\newblock In \emph{I}, 2023.

\bibitem[Carreira and Zisserman(2017)]{carreira2017quo}
Joao Carreira and Andrew Zisserman.
\newblock Quo vadis, action recognition? a new model and the kinetics dataset.
\newblock In \emph{proceedings of the IEEE Conference on Computer Vision and Pattern Recognition}, pages 6299--6308, 2017.

\bibitem[Chen et~al.(2020{\natexlab{a}})Chen, Frankle, Chang, Liu, Zhang, Wang, and Carbin]{chen2020lottery}
Tianlong Chen, Jonathan Frankle, Shiyu Chang, Sijia Liu, Yang Zhang, Zhangyang Wang, and Michael Carbin.
\newblock The lottery ticket hypothesis for pre-trained bert networks.
\newblock \emph{Advances in neural information processing systems}, 33:\penalty0 15834--15846, 2020{\natexlab{a}}.

\bibitem[Chen et~al.(2020{\natexlab{b}})Chen, Zhang, Liu, Chang, and Wang]{chen2020earlybert}
Tianlong Chen, Zhenyu Zhang, Sijia Liu, Shiyu Chang, and Zhangyang Wang.
\newblock Long live the lottery: The existence of winning tickets in lifelong learning.
\newblock In \emph{International Conference on Learning Representations}, 2020{\natexlab{b}}.

\bibitem[Chen et~al.(2021{\natexlab{a}})Chen, Cheng, Gan, Liu, and Wang]{chen2021ultra}
Tianlong Chen, Yu Cheng, Zhe Gan, Jingjing Liu, and Zhangyang Wang.
\newblock Ultra-data-efficient gan training: Drawing a lottery ticket first, then training it toughly.
\newblock \emph{arXiv preprint arXiv:2103.00397}, 3, 2021{\natexlab{a}}.

\bibitem[Chen et~al.(2021{\natexlab{b}})Chen, Sui, Chen, Zhang, and Wang]{chen2021unified}
Tianlong Chen, Yongduo Sui, Xuxi Chen, Aston Zhang, and Zhangyang Wang.
\newblock A unified lottery ticket hypothesis for graph neural networks.
\newblock In \emph{International conference on machine learning}, pages 1695--1706. PMLR, 2021{\natexlab{b}}.

\bibitem[Chen et~al.(2021{\natexlab{c}})Chen, Zhang, Yuan, Li, Deng, and Hu]{chen2021channel}
Yuxin Chen, Ziqi Zhang, Chunfeng Yuan, Bing Li, Ying Deng, and Weiming Hu.
\newblock Channel-wise topology refinement graph convolution for skeleton-based action recognition.
\newblock In \emph{Proceedings of the IEEE/CVF International Conference on Computer Vision}, pages 13359--13368, 2021{\natexlab{c}}.

\bibitem[Cheng et~al.(2020)Cheng, Zhang, He, Chen, Cheng, and Lu]{cheng2020skeleton}
Ke Cheng, Yifan Zhang, Xiangyu He, Weihan Chen, Jian Cheng, and Hanqing Lu.
\newblock Skeleton-based action recognition with shift graph convolutional network.
\newblock In \emph{Proceedings of the IEEE/CVF Conference on Computer Vision and Pattern Recognition}, pages 183--192, 2020.

\bibitem[Chijiwa et~al.(2021)Chijiwa, Yamaguchi, Ida, Umakoshi, and Inoue]{chijiwa2021pruning}
Daiki Chijiwa, Shin'ya Yamaguchi, Yasutoshi Ida, Kenji Umakoshi, and Tomohiro Inoue.
\newblock Pruning randomly initialized neural networks with iterative randomization.
\newblock \emph{Advances in Neural Information Processing Systems}, 34:\penalty0 4503--4513, 2021.

\bibitem[Das et~al.(2020)Das, Sharma, Dai, Bremond, and Thonnat]{das2020vpn}
Srijan Das, Saurav Sharma, Rui Dai, Francois Bremond, and Monique Thonnat.
\newblock Vpn: Learning video-pose embedding for activities of daily living.
\newblock In \emph{Computer Vision--ECCV 2020: 16th European Conference, Glasgow, UK, August 23--28, 2020, Proceedings, Part IX 16}, pages 72--90. Springer, 2020.

\bibitem[Ding et~al.(2018)Ding, He, Liu, and Liu]{ding2018combining}
Runwei Ding, Qinqin He, Hong Liu, and Mengyuan Liu.
\newblock Combining adaptive hierarchical depth motion maps with skeletal joints for human action recognition.
\newblock \emph{IEEE Access}, 7:\penalty0 5597--5608, 2018.

\bibitem[Ding et~al.(2023)Ding, Wen, Liu, Dai, Meng, and Liu]{ding2023integrating}
Runwei Ding, Yuhang Wen, Jinfu Liu, Nan Dai, Fanyang Meng, and Mengyuan Liu.
\newblock Integrating human parsing and pose network for human action recognition.
\newblock In \emph{CAAI International Conference on Artificial Intelligence}, pages 182--194. Springer, 2023.

\bibitem[Du et~al.(2015)Du, Wang, and Wang]{du2015hierarchical}
Yong Du, Wei Wang, and Liang Wang.
\newblock Hierarchical recurrent neural network for skeleton-based action recognition.
\newblock In \emph{Proceedings of the IEEE conference on computer vision and pattern recognition}, pages 1110--1118, 2015.

\bibitem[Duan et~al.(2022{\natexlab{a}})Duan, Wang, Chen, and Lin]{St-gcn++}
Haodong Duan, Jiaqi Wang, Kai Chen, and Dahua Lin.
\newblock Pyskl: Towards good practices for skeleton action recognition, 2022{\natexlab{a}}.

\bibitem[Duan et~al.(2022{\natexlab{b}})Duan, Wang, Chen, and Lin]{dg2022}
Haodong Duan, Jiaqi Wang, Kai Chen, and Dahua Lin.
\newblock Dg-stgcn: Dynamic spatial-temporal modeling for skeleton-based action recognition, 2022{\natexlab{b}}.

\bibitem[Duan et~al.(2022{\natexlab{c}})Duan, Wang, Chen, and Lin]{duan2022dg}
Haodong Duan, Jiaqi Wang, Kai Chen, and Dahua Lin.
\newblock Dg-stgcn: dynamic spatial-temporal modeling for skeleton-based action recognition.
\newblock \emph{arXiv preprint arXiv:2210.05895}, 2022{\natexlab{c}}.

\bibitem[Duan et~al.(2022{\natexlab{d}})Duan, Wang, Chen, and Lin]{duan2022pyskl}
Haodong Duan, Jiaqi Wang, Kai Chen, and Dahua Lin.
\newblock Pyskl: Towards good practices for skeleton action recognition, 2022{\natexlab{d}}.

\bibitem[Duan et~al.(2022{\natexlab{e}})Duan, Zhao, Chen, Lin, and Dai]{duan2022revisiting}
Haodong Duan, Yue Zhao, Kai Chen, Dahua Lin, and Bo Dai.
\newblock Revisiting skeleton-based action recognition.
\newblock In \emph{Proceedings of the IEEE/CVF Conference on Computer Vision and Pattern Recognition}, pages 2969--2978, 2022{\natexlab{e}}.

\bibitem[Evci et~al.(2020)Evci, Gale, Menick, Castro, and Elsen]{evci2020rigging}
Utku Evci, Trevor Gale, Jacob Menick, Pablo~Samuel Castro, and Erich Elsen.
\newblock Rigging the lottery: Making all tickets winners.
\newblock In \emph{International Conference on Machine Learning}, pages 2943--2952. PMLR, 2020.

\bibitem[Frankle and Carbin(2018)]{frankle2018lottery}
Jonathan Frankle and Michael Carbin.
\newblock The lottery ticket hypothesis: Finding sparse, trainable neural networks.
\newblock \emph{arXiv preprint arXiv:1803.03635}, 2018.

\bibitem[Frankle et~al.(2020)Frankle, Dziugaite, Roy, and Carbin]{frankle2020linear}
Jonathan Frankle, Gintare~Karolina Dziugaite, Daniel Roy, and Michael Carbin.
\newblock Linear mode connectivity and the lottery ticket hypothesis.
\newblock In \emph{International Conference on Machine Learning}, pages 3259--3269. PMLR, 2020.

\bibitem[Gan et~al.(2022)Gan, Chen, Li, Chen, Cheng, Wang, Liu, Wang, and Liu]{gan2022playing}
Zhe Gan, Yen-Chun Chen, Linjie Li, Tianlong Chen, Yu Cheng, Shuohang Wang, Jingjing Liu, Lijuan Wang, and Zicheng Liu.
\newblock Playing lottery tickets with vision and language.
\newblock In \emph{Proceedings of the AAAI Conference on Artificial Intelligence}, pages 652--660, 2022.

\bibitem[Gaur et~al.(2011)Gaur, Zhu, Song, and Roy-Chowdhury]{6126548}
U. Gaur, Y. Zhu, B. Song, and A. Roy-Chowdhury.
\newblock A “string of feature graphs” model for recognition of complex activities in natural videos.
\newblock In \emph{2011 International Conference on Computer Vision}, pages 2595--2602, 2011.

\bibitem[Gui et~al.(2018)Gui, Zhang, Wang, Liang, Moura, and Veloso]{8594452}
Liang-Yan Gui, Kevin Zhang, Yu-Xiong Wang, Xiaodan Liang, José M.~F. Moura, and Manuela Veloso.
\newblock Teaching robots to predict human motion.
\newblock In \emph{2018 IEEE/RSJ International Conference on Intelligent Robots and Systems (IROS)}, pages 562--567, 2018.

\bibitem[Ke et~al.(2017)Ke, Bennamoun, An, Sohel, and Boussaid]{ke2017new}
Qiuhong Ke, Mohammed Bennamoun, Senjian An, Ferdous Sohel, and Farid Boussaid.
\newblock A new representation of skeleton sequences for 3d action recognition.
\newblock In \emph{Proceedings of the IEEE conference on computer vision and pattern recognition}, pages 3288--3297, 2017.

\bibitem[Kumar and Kumar(2023)]{kumar2023multi}
Rahul Kumar and Shailender Kumar.
\newblock Multi-view multi-modal approach based on 5s-cnn and bilstm using skeleton, depth and rgb data for human activity recognition.
\newblock \emph{Wireless Personal Communications}, 130\penalty0 (2):\penalty0 1141--1159, 2023.

\bibitem[Li et~al.(2019)Li, Chen, Chen, Zhang, Wang, and Tian]{li2019actional}
Maosen Li, Siheng Chen, Xu Chen, Ya Zhang, Yanfeng Wang, and Qi Tian.
\newblock Actional-structural graph convolutional networks for skeleton-based action recognition.
\newblock In \emph{Proceedings of the IEEE/CVF conference on computer vision and pattern recognition}, pages 3595--3603, 2019.

\bibitem[Liu et~al.(2017)Liu, Wang, Duan, Abdiyeva, and Kot]{liu2017skeleton}
Jun Liu, Gang Wang, Ling-Yu Duan, Kamila Abdiyeva, and Alex~C Kot.
\newblock Skeleton-based human action recognition with global context-aware attention lstm networks.
\newblock \emph{IEEE Transactions on Image Processing}, 27\penalty0 (4):\penalty0 1586--1599, 2017.

\bibitem[Liu et~al.(2019)Liu, Shahroudy, Perez, Wang, Duan, and Kot]{liu2019ntu}
Jun Liu, Amir Shahroudy, Mauricio Perez, Gang Wang, Ling-Yu Duan, and Alex~C Kot.
\newblock Ntu rgb+ d 120: A large-scale benchmark for 3d human activity understanding.
\newblock \emph{IEEE transactions on pattern analysis and machine intelligence}, 42\penalty0 (10):\penalty0 2684--2701, 2019.

\bibitem[Liu et~al.(2020)Liu, Zhang, Chen, Wang, and Ouyang]{liu2020disentangling}
Ziyu Liu, Hongwen Zhang, Zhenghao Chen, Zhiyong Wang, and Wanli Ouyang.
\newblock Disentangling and unifying graph convolutions for skeleton-based action recognition.
\newblock In \emph{Proceedings of the IEEE/CVF conference on computer vision and pattern recognition}, pages 143--152, 2020.

\bibitem[Lozano et~al.(2009)Lozano, Abe, Liu, and Rosset]{lozano2009grouped}
Aur{\'e}lie~C Lozano, Naoki Abe, Yan Liu, and Saharon Rosset.
\newblock Grouped graphical granger modeling for gene expression regulatory networks discovery.
\newblock \emph{Bioinformatics}, 25\penalty0 (12):\penalty0 i110--i118, 2009.

\bibitem[Mikler(2022)]{mikler2022comparing}
Szymon~Jakub Mikler.
\newblock Comparing rewinding and fine-tuning in neural network pruning.
\newblock In \emph{ML Reproducibility Challenge 2021 (Fall Edition)}, 2022.

\bibitem[Ramanujan et~al.(2020)Ramanujan, Wortsman, Kembhavi, Farhadi, and Rastegari]{ramanujan2020s}
Vivek Ramanujan, Mitchell Wortsman, Aniruddha Kembhavi, Ali Farhadi, and Mohammad Rastegari.
\newblock What's hidden in a randomly weighted neural network?
\newblock In \emph{Proceedings of the IEEE/CVF Conference on Computer Vision and Pattern Recognition}, pages 11893--11902, 2020.

\bibitem[Shahroudy et~al.(2016)Shahroudy, Liu, Ng, and Wang]{shahroudy2016ntu}
Amir Shahroudy, Jun Liu, Tian-Tsong Ng, and Gang Wang.
\newblock Ntu rgb+ d: A large scale dataset for 3d human activity analysis.
\newblock In \emph{Proceedings of the IEEE conference on computer vision and pattern recognition}, pages 1010--1019, 2016.

\bibitem[Shao et~al.(2020)Shao, Zhao, Dai, and Lin]{shao2020finegym}
Dian Shao, Yue Zhao, Bo Dai, and Dahua Lin.
\newblock Finegym: A hierarchical video dataset for fine-grained action understanding.
\newblock In \emph{Proceedings of the IEEE/CVF conference on computer vision and pattern recognition}, pages 2616--2625, 2020.

\bibitem[Shi et~al.(2019{\natexlab{a}})Shi, Zhang, Cheng, and Lu]{shi2019skeleton}
Lei Shi, Yifan Zhang, Jian Cheng, and Hanqing Lu.
\newblock Skeleton-based action recognition with directed graph neural networks.
\newblock In \emph{Proceedings of the IEEE/CVF conference on computer vision and pattern recognition}, pages 7912--7921, 2019{\natexlab{a}}.

\bibitem[Shi et~al.(2019{\natexlab{b}})Shi, Zhang, Cheng, and Lu]{shi2019two}
Lei Shi, Yifan Zhang, Jian Cheng, and Hanqing Lu.
\newblock Two-stream adaptive graph convolutional networks for skeleton-based action recognition.
\newblock In \emph{Proceedings of the IEEE/CVF conference on computer vision and pattern recognition}, pages 12026--12035, 2019{\natexlab{b}}.

\bibitem[Shi et~al.(2020{\natexlab{a}})Shi, Zhang, Cheng, and Lu]{shi2020decoupled}
Lei Shi, Yifan Zhang, Jian Cheng, and Hanqing Lu.
\newblock Decoupled spatial-temporal attention network for skeleton-based action-gesture recognition.
\newblock In \emph{Proceedings of the Asian Conference on Computer Vision}, 2020{\natexlab{a}}.

\bibitem[Shi et~al.(2020{\natexlab{b}})Shi, Zhang, Cheng, and Lu]{shi2020skeleton}
Lei Shi, Yifan Zhang, Jian Cheng, and Hanqing Lu.
\newblock Skeleton-based action recognition with multi-stream adaptive graph convolutional networks.
\newblock \emph{IEEE Transactions on Image Processing}, 29:\penalty0 9532--9545, 2020{\natexlab{b}}.

\bibitem[Si et~al.(2018)Si, Jing, Wang, Wang, and Tan]{si2018skeleton}
Chenyang Si, Ya Jing, Wei Wang, Liang Wang, and Tieniu Tan.
\newblock Skeleton-based action recognition with spatial reasoning and temporal stack learning.
\newblock In \emph{Proceedings of the European conference on computer vision (ECCV)}, pages 103--118, 2018.

\bibitem[Simonyan and Zisserman(2014)]{simonyan2014two}
Karen Simonyan and Andrew Zisserman.
\newblock Two-stream convolutional networks for action recognition in videos.
\newblock \emph{Advances in neural information processing systems}, 27, 2014.

\bibitem[Song et~al.(2020)Song, Zhang, Shan, and Wang]{song2020richly}
Yi-Fan Song, Zhang Zhang, Caifeng Shan, and Liang Wang.
\newblock Richly activated graph convolutional network for robust skeleton-based action recognition.
\newblock \emph{IEEE Transactions on Circuits and Systems for Video Technology}, 31\penalty0 (5):\penalty0 1915--1925, 2020.

\bibitem[Vemulapalli et~al.(2014)Vemulapalli, Arrate, and Chellappa]{vemulapalli2014human}
Raviteja Vemulapalli, Felipe Arrate, and Rama Chellappa.
\newblock Human action recognition by representing 3d skeletons as points in a lie group.
\newblock In \emph{Proceedings of the IEEE conference on computer vision and pattern recognition}, pages 588--595, 2014.

\bibitem[Verma et~al.(2020)Verma, Sah, and Srivastava]{verma2020deep}
Pratishtha Verma, Animesh Sah, and Rajeev Srivastava.
\newblock Deep learning-based multi-modal approach using rgb and skeleton sequences for human activity recognition.
\newblock \emph{Multimedia Systems}, 26\penalty0 (6):\penalty0 671--685, 2020.

\bibitem[Wang et~al.(2014)Wang, Liu, Wu, and Yuan]{6626306}
Jiang Wang, Zicheng Liu, Ying Wu, and Junsong Yuan.
\newblock Learning actionlet ensemble for 3d human action recognition.
\newblock \emph{IEEE Transactions on Pattern Analysis and Machine Intelligence}, 36\penalty0 (5):\penalty0 914--927, 2014.

\bibitem[Xie et~al.(2024{\natexlab{a}})Xie, Meng, Zhao, Anh, Yang, and Zheng]{xie2024dynamicTIP}
Jianyang Xie, Yanda Meng, Yitian Zhao, Nguyen Anh, Xiaoyun Yang, and Yalin Zheng.
\newblock Dynamic semantic-based spatial-temporal graph convolution network for skeleton-based human action recognition.
\newblock \emph{IEEE Transactions on Image Processing}, 2024{\natexlab{a}}.

\bibitem[Xie et~al.(2024{\natexlab{b}})Xie, Meng, Zhao, Nguyen, Yang, and Zheng]{xie2024dynamic}
Jianyang Xie, Yanda Meng, Yitian Zhao, Anh Nguyen, Xiaoyun Yang, and Yalin Zheng.
\newblock Dynamic semantic-based spatial graph convolution network for skeleton-based human action recognition.
\newblock In \emph{Proceedings of the AAAI Conference on Artificial Intelligence}, pages 6225--6233, 2024{\natexlab{b}}.

\bibitem[Yan et~al.(2018)Yan, Xiong, and Lin]{yan2018spatial}
Sijie Yan, Yuanjun Xiong, and Dahua Lin.
\newblock Spatial temporal graph convolutional networks for skeleton-based action recognition.
\newblock In \emph{Thirty-second AAAI conference on artificial intelligence}, 2018.

\bibitem[Yang et~al.(2021)Yang, Yan, Zhang, Sun, Li, and Maybank]{yang2021feedback}
Hao Yang, Dan Yan, Li Zhang, Yunda Sun, Dong Li, and Stephen~J Maybank.
\newblock Feedback graph convolutional network for skeleton-based action recognition.
\newblock \emph{IEEE Transactions on Image Processing}, 31:\penalty0 164--175, 2021.

\bibitem[Zhang et~al.(2020)Zhang, Lan, Zeng, Xing, Xue, and Zheng]{zhang2020semantics}
Pengfei Zhang, Cuiling Lan, Wenjun Zeng, Junliang Xing, Jianru Xue, and Nanning Zheng.
\newblock Semantics-guided neural networks for efficient skeleton-based human action recognition.
\newblock In \emph{proceedings of the IEEE/CVF conference on computer vision and pattern recognition}, pages 1112--1121, 2020.

\bibitem[Zhang et~al.(2017)Zhang, Liu, and Xiao]{zhang2017geometric}
Songyang Zhang, Xiaoming Liu, and Jun Xiao.
\newblock On geometric features for skeleton-based action recognition using multilayer lstm networks.
\newblock In \emph{2017 IEEE Winter Conference on Applications of Computer Vision (WACV)}, pages 148--157. IEEE, 2017.

\bibitem[Zhou et~al.(2019)Zhou, Lan, Liu, and Yosinski]{zhou2019deconstructing}
Hattie Zhou, Janice Lan, Rosanne Liu, and Jason Yosinski.
\newblock Deconstructing lottery tickets: Zeros, signs, and the supermask.
\newblock \emph{Advances in neural information processing systems}, 32, 2019.

\end{thebibliography}
}


\end{document}


\maketitle
\section{Appendix / supplemental material}
\subsection{Implementations Details}
\label{implementation}
All experiments are conducted on one A100 GPU with the PyTorch deep learning framework. All models are trained for 100 epochs with the Cosine Annealing learning rate scheduler by using SGD with momentum 0.9, weight decay $5e^{-4}$. The initial learning rate was set to 0.1. The batch size was set to 128. To accelerate the training process, the input of temporal length was set to 64 in the ablation study. For a fair comparison, the input of temporal length was set to 100 when comparing the stare-of-the-arts. The pre-processing approach follows the setting in~\cite{St-gcn++}.
\subsection{Datasets}
\label{dataset}

\textbf{NTU RGB+D 60}~\cite{shahroudy2016ntu}. The action samples are performed by 40 volunteers and categorized into 60 classes. Each sample contains an action and is guaranteed to have at most 2 subjects, which are captured by three Microsoft Kinect v2 cameras from different views concurrently. Two benchmarksre were commend : (1) cross-subject (NTU60-Xsub): training data comes from 20 subjects, and testing data comes from the other 20 subjects. (2) cross-view (NTU60--view): training data comes from camera views 2 and 3, and testing data comes from camera view 1. 

\textbf{NTU RGB+D 120}~\cite{liu2019ntu}. NTU RGB+D 120 is currently the largest dataset with 3D joint annotations for HAR, which extends NTU RGB+D 60 with additional 57,367 skeleton sequences over 60 extra action classes. Totally 113,945 samples over 120 classes are performed by 106 volunteers, captured with three camera views. This dataset contains 32 setups, each denoting a specific location and background. The authors of this dataset recommend two benchmarks: (1) cross-subject (NTU120-Xsub): training data comes from 53 subjects, and testing data comes from the other 53 subjects. (2) cross-setup (NTU120-Xset): training data comes from samples with even setup IDs, and testing data comes from samples with odd setup IDs.

\textbf{Kinetics-400}~\cite{carreira2017quo}. Kinetics-400 is a large-scale action recognition dataset with 400 actions. The skeletons were provided by~\cite{St-gcn++}, where the Openose algorithm~\cite{cao2017realtime} was applied for joint estimation. The box threshold of human detection is set as 0.5. After the validation, there are a total of 236,489 skeleton sequences for training and 19,505 skeleton sequences for testing.

\textbf{FineGYM}~\cite{shao2020finegym}. FineGYM is a fine-grained action recognition dataset with 29,000 videos of 99 fine-grained gymnastic action classes. Skeletons are extracted with ground-truth human bounding boxes as described in~\cite{duan2022revisiting}.
{
    \small
    \bibliographystyle{ieeenat_fullname}
    \bibliography{main}
}
